\documentclass[pdflatex,sn-mathphys-num]{sn-jnl}

\usepackage{graphicx}
\usepackage{multirow}
\usepackage{amsmath,amssymb,amsfonts}
\usepackage{amsthm}
\usepackage{mathrsfs}
\usepackage[title]{appendix}
\usepackage[table]{xcolor}
\usepackage{textcomp}
\usepackage{manyfoot}
\usepackage{booktabs}
\usepackage{listings}
\usepackage{algorithm}
\usepackage{algpseudocode}

\theoremstyle{thmstyleone}%
\theoremstyle{thmstyletwo}%

\theoremstyle{thmstylethree}%

\begin{document}

\title[Article Title]{A Meta-Knowledge–Augmented LLM Framework for Hyperparameter Optimization in Time-Series Forecasting}


\author*[1]{\fnm{Ons} \sur{Saadallah}}\email{saadallahons@gmail.com}

\author[1,2]{\fnm{Mátyás} \sur{Andó}}\email{am@inf.elte.hu}

\author[1]{\fnm{Tamás Gábor} \sur{Orosz}}\email{orosztamas@inf.elte.hu}

\affil*[1]{\orgdiv{Department of Data Science and Technology}, \orgname{Eötvös Loránd University}, \orgaddress{\city{Budapest}, \country{Hungary}}}

\affil[2]{\orgdiv{Department of Data Science and Technology, Savaria Institute of Technology, Faculty of Informatics}, \orgname{Eötvös Loránd University}, \orgaddress{\city{Budapest}, \country{Hungary}}}


\abstract{Hyperparameter optimization (HPO) plays a central role in the performance of deep learning models, yet remains computationally expensive and difficult to interpret, particularly for time-series forecasting. Although Bayesian Optimization (BO) is a standard approach, it typically treats tuning tasks independently and provides limited insight into its decisions. Recent advances in large language models (LLMs) offer new opportunities for incorporating structured prior knowledge and reasoning into optimization pipelines.
We introduce LLM-AutoOpt, a hybrid HPO framework that combines BO with LLM-based contextual reasoning to address these issues. The framework encodes dataset meta-features, model descriptions, historical optimization outcomes, and target objectives as structured meta-knowledge within LLM prompts, using BO to initialize the search and mitigate the cold-start effects. This design enables context-aware and stable hyperparameter refinement while exposing the reasoning behind the optimization decisions.
Experiments on a multivariate time series forecasting benchmark demonstrate that LLM-AutoOpt achieves improved predictive performance and more interpretable optimization behavior compared to BO and LLM baselines without meta-knowledge.}




\maketitle

\section{Introduction}\label{sec1}
Auto Ml is the process of automating the tasks involved in building and deploying machine learning models~\citep{ibm_automl_2025}. Hyperparameter optimization (HPO) is one of the most challenging components of AutoML, and is essential for developing robust and effective deep learning models. The goal of HPO is to identify optimal configurations that enhance the model performance, stability, and interpretability. Traditional HPO methods, such as grid search, random search, Bayesian optimization, and gradient-based approaches, are commonly used to optimize hyperparameters and have demonstrated improvements in model accuracy ~\cite{bib1}. However, these methods still struggle with large-scale data and complex models, including deep neural networks, and are often computationally expensive ~\cite{bib2,bib3,bib4}.AutoML enhances the hyperparameter optimization task by incorporating different learning paradigms such as meta-learning and reinforcement learning.

Meta-learning enables hyperparameter optimization (HPO) methods to rapidly
adapt to new tasks by leveraging experience from previously observed tasks, where each task is represented through informative meta-features ~\cite{bib5}. By transferring prior optimization knowledge, meta-learning–based HPO significantly improves both sample efficiency and adaptability compared to conventional approaches that treat each task independently. Crucially, these methods mitigate the need to initialize the optimization process from scratch, instead warm-starting the search using insights derived from historical tasks ~\cite{bib6,bib7,bib8,bib9}.This paradigm has been successfully operationalized in several AutoML systems, most notablyAuto-sklearn ~\cite{bib11}, demonstrates its practical effectiveness in real-world settings.

As large language models (LLMs) have become increasingly integrated into machine learning research, they have also been applied to hyperparameter optimization (HPO). Prior work has leveraged LLMs to automate tasks such as neural architecture search and end-to-end ML pipeline construction ~\cite{bib12,bib13}.Owing to their strong in-context learning capabilities, LLMs can rapidly adapt to new tasks without explicit fine-tuning~\cite{bib14,bib15}.This property has recently been extended to optimization settings, where LLMs are explored as alternatives or complements to traditional HPO methods. For example, Bayesian optimization
has been reformulated as a natural language reasoning task, enabling LLMs to exploit their contextual reasoning abilities to recommend improved hyperparameter configurations~\cite{bib16}.More recent approaches, such as MetaLLMix ~\cite{bib17}, integrate meta-learning with LLM reasoning and explainable AI to transfer knowledge across tasks using historical experiment meta-datasets. In this framework, LLMs reason over structured meta-information to generate task-specific model and hyperparameter recommendations, with SHAP-based explanations enhancing transparency. While such methods have shown promising results for classification and regression tasks~\cite{bib18}, their applicability to time-series forecasting remains largely unexplored due to the presence of strong temporal dependencies and evolving data distributions. Prior work has operationalized meta-learning for hyperparameter optimization through several complementary strategies. Feurer et al.~\cite{bib9} demonstrate that Bayesian optimization can be effectively warm-started by conditioning surrogate models on dataset meta-features, thereby guiding the search toward promising regions of the hyperparameter space. In parallel, In ~\cite{bib8}, the authors propose transferring high-performing configurations from metadatasets of previous tuning runs to initialize optimization on new tasks. Extending these ideas, Lindauer and Hutter ~\cite{bib10} show that incorporating historical benchmark data into warm-started, model-based configuration frameworks improves scalability while preserving optimization efficiency.Collectively, these methods
highlight how meta-knowledge can be systematically exploited to accelerate
convergence and improve solution quality relative to cold-start optimization.

To address this, we propose LLM-AutoMOpt, a framework for automating
hyperparameter optimization in deep learning models for time series forecasting.The framework follows an iterative optimization process in which Bayesian optimization is first employed to initialize the search space and provide informative prior configurations, which are subsequently refined through structured LLMguided reasoning. The LLM operates on carefully constructed prompts enriched
with meta-knowledge, including dataset characteristics, model descriptions, target loss objectives, and historical optimization outcomes. This structured metainformation enables the model to generate context-aware and performance-driven
hyperparameter recommendations. A key challenge lies in prompt design, which must balance expressive reasoning with well-defined constraints to mitigate hallucination and ensure stable optimization behavior. Beyond identifying optimal configurations, LLM-AutoMOpt emphasizes transparency and interpretability, explicitly exposing the meta-knowledge and reasoning processes underlying its optimization decisions.

\section{Related Works}\label{sec2}
Hyperparameter optimization (HPO) seeks to automatically identify hyperparameter configurations that maximize a model’s performance on a given task, often at the cost of expensive and time-consuming model evaluations~\citep{ibm_automl_2025}.Early approaches such as grid search systematically enumerate predefined configurations, guaranteeing coverage of the search space but becoming computationally prohibitive as the number or range of hyperparameters grows~\cite{bib19}.Random search addresses this limitation by sampling configurations independently, frequently achieving superior efficiency, particularly when only a small subset of hyperparameters has a dominant effect on performance~\cite{bib20}. Bayesian optimization further advances HPO by learning a probabilistic surrogate of the objective function and selecting new configurations via an explicit exploration–exploitation trade-off, enabling faster convergence with significantly fewer evaluations~\cite{bib21}. Despite their success, these methods generally treat each optimization problem independently, failing to exploit knowledge from prior tasks ~\cite{bib2}. This limitation reduces scalability and sample efficiency, especially in computationally demanding settings such as time-series forecasting, where training costs are high and model behavior is strongly data-dependent~\cite{bib3}.

To overcome this limitation, meta-learning has emerged as an effective
paradigm for improving scalability and resource efficiency across a wide range of learning problems~\cite{bib22}. By leveraging knowledge acquired from previously solved tasks, meta-learning enables hyperparameter optimization to adapt more rapidly to new tasks, effectively avoiding repeated optimization from scratch. In particular, Bayesian optimization has been shown to benefit significantly from meta-learning–driven warm starts, where prior task information guides the initial search process ~\cite{bib23,bib24}.This combination has been successfully applied to complex attention-based architectures,leading to improved forecasting accuracy in energy systems compared to standalone optimization approaches ~\cite{bib25}.Building on this idea, meta-Bayesian optimization learns informative surrogate model priors from related tasks, further accelerating convergence and improving optimization efficiency on new problems ~\cite{bib26}.

Beyond Bayesian and meta-learning–based approaches, reinforcement learning
(RL) has also been explored for hyperparameter optimization by formulating
HPO as a sequential decision-making problem. In this setting, an RL agent
iteratively selects hyperparameter configurations based on feedback from previous evaluations, learning a policy that aims to maximize expected performance while reducing the number of costly evaluations ~\cite{bib27}. Building on this idea, metalearning has been combined with RL to further improve efficiency. For example, MultiTaskPB2 leverages metadata from prior RL training runs to warm-start hyperparameter optimization and bias surrogate models for new tasks, achieving significantly faster and more effective optimization than standard methods ~\cite{bib28}.

More recently, large language models (LLMs) have introduced a new paradigm
for hyperparameter optimization by leveraging natural language understanding,contextual reasoning, and few-shot generalization capabilities s~\cite{bib29,bib30,bib32,bib33}. In AgentHPO ~\cite{bib29}, an LLM-based agent autonomously analyzes task descriptions and historical trials, proposes new hyperparameter configurations, executes experiments, and iteratively refines its recommendations, forming a human-like
and interpretable optimization loop that reduces manual intervention. LLMs
have also been integrated with traditional HPO strategies to improve efficiency.For instance, coarse grid search can be guided by an LLM to identify and refine promising regions of the search space, effectively pruning unlikely configurations and reducing computational cost while maintaining or improving tuning quality ~\cite{bib32}. Furthermore, LLMs have been combined with Bayesian optimization in frameworks such as LLAMBO, where the BO process is reformulated as a natural language reasoning task. This enables LLMs to propose and evaluate candidate
configurations conditioned on past trials, supporting zero-shot warm-starting, improved surrogate modeling, and more focused exploration of promising regions, ultimately leading to faster convergence and superior hyperparameter tuning compared to conventional BO~\cite{bib16}.Beyond directly steering the optimizationprocess, recent research has investigated how LLMs can be used as meta-learners that reason over collections of prior experiments. Frameworks such as MetaLLMix
~\cite{bib17} construct structured meta-databases from historical training runs and rely on LLMs to synthesize transferable optimization knowledge across tasks. Related in-context meta-learning approaches ~\cite{bib18} encode datasets and experimental settings as structured, human-readable metadata, enabling LLMs to infer suitable hyperparameter configurations in zero-shot or few-shot regimes.

Unlike agent-based or BO-integrated LLM methods, these approaches emphasize knowledge transfer at the task level rather than iterative trial-by-trial optimization. However, they have primarily been evaluated on classification and regression benchmarks and do not explicitly address the domain-specific challenges of time-series forecasting, where strong temporal dependencies, heterogeneous sampling rates, seasonal dynamics, and high training costs require specialized meta-features and optimization strategies. To address these limitations, we propose LLM-AutoMOpt, a forecasting-specific HPO framework that tightly integrates Bayesian optimization for structured initialization with LLM based in-context reasoning over rich temporal meta-knowledge. In contrast to prior work, LLM-AutoMOpt explicitly accounts for temporal structure and model dynamics, enabling more sample-efficient, interpretable, and adaptive hyperparameter optimization for deep sequence models such as LSTM, Bi-LSTM, and Transformer architectures.
\section{Materials and Methods}

This section presents the methodology we designed to evaluate the capabilities of \textbf{LLM-AutoOpt} in performing hyperparameter optimization for deep learning models in time-series forecasting. Our goal is to assess how effectively large language models (LLMs) can recommend high-quality hyperparameter configurations while providing interpretable reasoning and leveraging task-specific meta-knowledge. To this end, we first formalize the \textit{problem statement} in Sect.~\ref{sec:problem_statement}. We then describe the construction of forecasting-specific meta-knowledge in Sect.~\ref{sec:meta_knowledge}, followed by the structured prompt design in Sect.~\ref{sec:prompt_design}, and finally, the overall LLM-AutoOpt optimization workflow in Sect.~\ref{sec:workflow}.
\subsection{Problem Statement}
\label{sec:problem_statement}

We consider the problem of hyperparameter optimization (HPO) for deep learning models applied to time-series forecasting. Let $\mathcal{D} = \{(x_t, y_t)\}_{t=1}^T$ denote a univariate or multivariate time-series dataset with $T$ sequential observations, and let $\mathcal{M}_\theta$ represent a forecasting model parameterized by weights $\theta$. The model's behavior is controlled by a set of hyperparameters $\lambda$ drawn from a search space $\Lambda$, and its predictive performance is evaluated using a loss function $\mathcal{L}(\mathcal{M}_\theta, \mathcal{D})$.

The HPO task can be formally expressed as the search for the optimal hyperparameter configuration $\lambda^*$ that minimizes the expected validation loss:

\begin{equation}
\lambda^* = \arg \min_{\lambda \in \Lambda} \ \mathbb{E}_{\theta \sim \text{Train}(\mathcal{M}_\lambda, \mathcal{D})} \Big[ \mathcal{L}(\mathcal{M}_\theta, \mathcal{D}_{\text{val}}) \Big],
\end{equation}

where $\mathcal{M}_\lambda$ denotes the model trained under hyperparameters $\lambda$, $\mathcal{D}_{\text{val}}$ is a validation set, and $\text{Train}(\mathcal{M}_\lambda, \mathcal{D})$ represents the stochastic training procedure that produces model parameters $\theta$.

The primary objective is to efficiently identify $\lambda^*$ while providing interpretable insights into the search process. This formulation naturally accounts for the stochastic nature of model training and the high-dimensional, often complex structure of the hyperparameter space, which poses significant challenges in both optimization efficiency and result interpretability.
\subsection{Forecasting-Specific Meta-Knowledge}
\label{sec:meta_knowledge}
Meta-knowledge provides the contextual foundation for LLM-AutoOpt, enabling the LLM to reason over task characteristics, prior optimization behavior, and modeling constraints. Unlike conventional HPO methods that rely solely on objective evaluations, LLM-AutoOpt conditions optimization on structured information describing the forecasting problem and the optimization process itself.
\subsection{Per-Feature Meta-Feature Extraction}
\label{sec:meta_features}
We extract a comprehensive set of meta-features to capture the statistical, temporal, and structural properties of the input timeseries. These include descriptive statistics (e.g., mean, variance, percentiles), distributional measures (skewness, kurtosis), and variability indicators (range, interquartile range, coefficient of variation). Temporal dependencies are quantified using autocorrelation and partial autocorrelation coefficients at multiple lags, while trend and seasonality strength are estimated via decomposition-based measures. Additional features describe data quality and complexity, including missing value ratios, outlier proportions, peak and trough counts, nonlinearity proxies, and stationarity diagnostics such as the Augmented Dickey–Fuller (ADF)  statistic and p-value. Together, these meta-features provide a compact yet expressive representation of the forecasting task, allowing the LLM to adapt its recommendations to diverse temporal regimes.
\subsubsection{Meta-Feature Statistical Summarization}
Beyond providing raw meta-features, we derive higher-level statistical summaries that aggregate these descriptors into interpretable indicators of the dataset's behavior. Examples include summarizing autocorrelation coefficients across multiple lags to characterize overall temporal dependence, aggregating trend and stationarity statistics (e.g., ADF statistics and trend slopes) to describe global non-stationarity, and combining volatility and noise-related measures (e.g., coefficient of variation, residual variance, and outlier ratio) to estimate the effective noise level of the series. These summarized representations are injected into the prompt in human-readable form, enabling the LLM to reason over dataset characteristics at a semantic level rather than directly interpreting raw numerical statistics. This Abstraction facilitates better generalization in decision-making and significantly reduces hallucinated reasoning by constraining the LLM’s inference process to validate dataset-derived indicators.
\subsubsection{Bayesian Optimization Initialization}
We employ Bayesian Optimization (BO) to efficiently explore the hyperparameter space. Let $\lambda_{\text{BO}}^{\min}$ denote the best configuration found:
\begin{equation}
\lambda_{\text{BO}}^{\min} = \arg\min_{\lambda \in \Lambda} \mathcal{L}(\mathcal{M}_\lambda, \mathcal{D}_{\text{val}})
\end{equation}
This configuration is provided as meta-knowledge to the LLM, giving it a strong inductive bias.
\subsubsection{Model Specification}
The model specification explicitly encodes the architecture
and training setup of the candidate forecasting model, including layer types, hidden dimensions, activation functions, optimizers, learning rates, and loss functions. Providing this information as structured meta-knowledge grounds the LLM’s reasoning in the intended experimental design, reducing hallucination and ensuring that proposed hyperparameter updates remain consistent with the
model’s functional constraints.
\subsubsection{Search Space Constraints}
The hyperparameter search space is defined through
explicit bounds and discrete choices for each parameter. These constraints serve both as a safety mechanism and as a reasoning aid, preventing invalid configurations (e.g., negative learning rates or unsupported batch sizes) while narrowing the space of plausible recommendations. Explicit constraints are essential for stable interaction between the LLM and the training pipeline. It is designed to act as a trust region centered around the current best-performing configuration.This formulation aims to reduce variance in the LLM’s recommendations while preserving controlled exploration of promising hyperparameter regions.
\subsubsection{Few-Shot Context (Optional)}
Representative input-output examples may be included to provide additional grounding, allowing the LLM to reason about scale, variability, and noise characteristics.
\subsubsection{Target Loss Criterion}
The iterative optimization stops once the target loss threshold is achieved:
\begin{equation}
\mathcal{L}_{\text{target}} = \mathcal{L}_{\text{BO}}^{\min} - \epsilon, \quad \epsilon > 0
\end{equation}

\subsection{Structured Prompt Design}
\label{sec:prompt_design}
The meta-knowledge is embedded within a structured framework, a zero-shot prompt that conditions the LLM to generate hyperparameter updates. The prompt is designed to balance expressive reasoning with strict structural constraints. It enforces a single valid JSON output, prohibits duplicate hyperparameters, and requires adherence to predefined types and search space bounds. Internal reasoning fields encourage the LLM to justify each recommendation based on the provided meta-knowledge, improving transparency without exposing free-form text that could destabilize downstream processing. This design minimizes hallucinations, ensures consistency across iterations, and enables reliable integration of LLM outputs into the optimization loop. This ensures interpretability, minimizes hallucinations, and guarantees consistency for downstream training.
\subsection{LLM-AutoOpt Optimization Workflow}
\label{sec:workflow}

\begin{figure}[!htbp]
    \centering
    \includegraphics[width=0.5\textwidth]{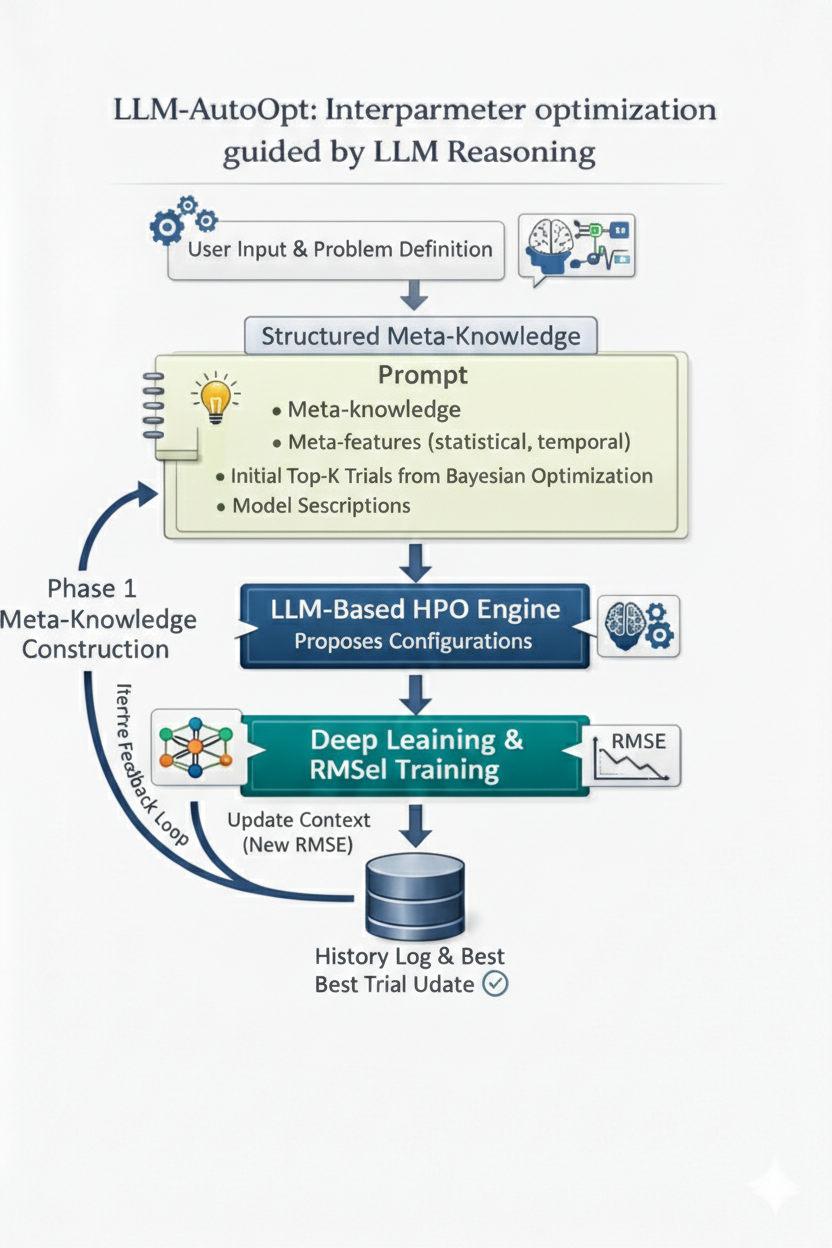}
    \caption{\large Automated hyperparameter optimization workflow guided by LLM reasoning.}
    \label{fig:Fig1}
\end{figure}

The workflow consists of two main phases as shown in Figure \ref{fig:Fig1}:

\paragraph{Phase 1: Meta-Knowledge Construction}
Meta-features are first extracted from the input time-series data to characterize its temporal and statistical structure. Bayesian Optimization is then applied to explore the hyperparameter space of the candidate model. The best-performing trial, together with the extracted meta-features, model specification, and search space constraints, forms the initial meta-knowledge base.

\paragraph{Phase 2: LLM-Guided Iterative Refinement}
The meta-knowledge is injected into the structured prompt and provided to the LLM, which proposes refined hyperparameter configurations. The candidate model is trained using these recommendations, and the resulting performance metric (e.g., RMSE) is fed back into the prompt as updated context. This iterative loop continues until the target loss criterion is satisfied. The number of initial trials is dynamically increased whenever a hyperparameter configuration outperforms the current best RMSE identified by Bayesian optimization.

\noindent
\textbf{Advantages:} The framework is sample-efficient, interpretable, task-aware, and provides iterative refinement that adapts to historical optimization behavior and dataset characteristics.

\section{Experimental Evaluation}\label{sec4}
This section evaluates the effectiveness of LLM-guided hyperparameter optimization, with a focus on both predictive performance and interpretability. In particular, we investigate how structured meta-knowledge influences optimization efficiency, model accuracy, and the transparency of the decision-making process. Our evaluation is guided by the following research questions:
\begin{enumerate}
    \item \textbf{RQ1 (Performance):} How does the proposed LLM-AutoOpt framework compare to Bayesian Optimization in terms of forecasting accuracy (RMSE)?
    \item \textbf{RQ2 (Efficiency):} How do training time and total optimization time differ between LLM-based HPO and BO-based baselines?
    \item \textbf{RQ3 (Optimization Dynamics):} Does LLM-guided refinement converge to high-quality hyperparameter configurations with fewer evaluations than BO?
    \item \textbf{RQ4 (Role of Meta-Knowledge):} How does incorporating forecasting-specific meta-knowledge affect the quality and stability of LLM-generated hyperparameter recommendations?
    \item \textbf{RQ5 (Interpretability):} To what extent does LLM-AutoOpt improve interpretability by exposing reasoning signals and meta-knowledge usage compared to black-box optimization methods?
\end{enumerate}
\subsection{Experimental Setup}

\subsubsection{Dataset and Preprocessing}
We conduct experiments on the \textit{Jena Climate} multivariate time-series dataset, collected at the meteorological station of the Max Planck Institute for Biogeochemistry in Jena, Germany. The dataset contains high-frequency measurements recorded every 10 minutes between January 1, 2004 and December 31, 2020 and is widely used as a benchmark for multivariate forecasting tasks \cite{bib34}. Each timestamp includes 14 meteorological variables such as air temperature, humidity, atmospheric pressure, and wind characteristics.To ensure computational tractability while preserving temporal structure, we extract a continuous six-month subset of the data. The CO\(_2\) variable is excluded due to extensive missing values. The remaining variables, air temperature (T), relative humidity (rh), atmospheric pressure (p), and wind speed (wv), are selected due to their relevance and complementary physical interpretations.

All variables are normalized using min–max scaling. The forecasting horizon is set to four time stamps, i.e., 40 minutes ahead. The data are split chronologically into training (70\%) and testing (30\%) sets to prevent information leakage across time.
\subsubsection{Forecasting Model}
\paragraph{Bidirectional LSTM (Bi-LSTM)}
We select the Bidirectional Long Short-Term Memory (Bi-LSTM) network as the candidate forecasting model due to its widespread use and effectiveness in multivariate time-series modeling 
\cite{bib35}. Bi-LSTMs process sequences in both forward and backward directions, enabling the capture of long-range temporal dependencies. 
\paragraph{Optimized Hyperparameters}
We optimize a set of hyperparameters that strongly influence model performance and generalization, including batch size, hidden dimension, number of layers, dropout rate, learning rate, number of epochs, lag length, and optimizer choice. Table~\ref{tab:hyperparameters} summarizes these parameters and their effects. A randomly initialized baseline configuration (Table~\ref{tab:baseline}) is used for comparison.

\begin{table}[h!]
\centering
\caption{List of Optimized Hyperparameters}
\label{tab:hyperparameters}
\rowcolors{2}{gray!10}{white} 
\begin{tabular}{p{3cm} p{10cm}}
\hline
\rowcolor{gray!30} 
\textbf{Hyperparameter} & \textbf{Description \& Impact} \\
\hline
Batch size & Number of samples per update; affects \textbf{training stability} and \textbf{generalization}. \\
Hidden size & Number of LSTM units; larger size captures \textbf{complex patterns} but may overfit. \\
Number of layers & Stacked LSTM layers; more layers capture \textbf{long-term dependencies} but may overfit. \\
Dropout & Fraction of neurons ignored during training; prevents \textbf{overfitting}, improves \textbf{robustness}. \\
Learning rate (lr) & Step size for weight updates; high lr speeds up training but may be \textbf{unstable}. \\
Epochs & Number of passes over data; more epochs improve learning but may \textbf{overfit}. \\
Lag & Number of previous time steps used as input; captures \textbf{temporal dependencies}. \\
Optimizer & Method for updating weights (e.g., Adam, SGD); affects \textbf{convergence} and \textbf{accuracy}. \\
\hline
\end{tabular}
\end{table}
\begin{table}[h!]
\centering
\caption{Baseline Hyperparameters for the Bi-LSTM Model}
\label{tab:baseline}
\rowcolors{2}{gray!10}{white}
\begin{tabular}{p{5cm} p{4cm}}
\hline
\rowcolor{gray!30}
\textbf{Hyperparameter} & \textbf{Value} \\
\hline
Hidden size & 5 \\
Number of layers & 6 \\
Dropout & 0.5 \\
Lag & 3 \\
Batch size & 128 \\
Epochs & 10 \\
Learning rate & $1 \times 10^{-9}$ \\
Optimizer & Adam \\
\hline
\end{tabular}
\end{table}
\subsubsection{Large Language Model Backbone}
The LLM backbone used for hyperparameter reasoning is Qwen2.5-72B, a large-scale language model developed by
the Qwen team at Alibaba Group. The model is pretrained on multilingual and
multimodal data and further aligned using high-quality instruction datasets.
Its strong contextual reasoning capabilities make it well-suited for structured
decision-making tasks such as hyperparameter optimization. Prior studies show
that Qwen2.5-72B achieves low regret as the number of support tasks increases,
outperforming alternative LLMs in optimization-oriented settings \cite{bib17,bib18}. All
LLM outputs are validated against predefined constraints to ensure correctness
and feasibility. To balance deterministic behavior and mitigate hallucinations,
the decoding temperature parameter was set to 0.2.
\subsubsection{Meta-Knowledge Prompt Configuration}
The LLM prompt integrates structured meta-knowledge from all stages of the ML pipeline:
\paragraph{Data Meta-Knowledge}
Statistical, temporal, and structural meta-features describing the multivariate time series are provided (Table~\ref{tab:meta_features}). These features summarize
distributional properties, temporal dependencies, seasonality, stationarity, and
data quality.
\begin{table}[h!]
\centering
\small
\caption{Meta-feature definitions and formulas used to summarize statistical and temporal properties of a univariate time series $\{x_t\}_{t=1}^T$.}
\label{tab:meta_features}
\rowcolors{2}{gray!10}{white}
\begin{tabular}{p{3cm} p{9cm}}
\toprule
\textbf{Meta-feature} & \textbf{Definition / Formula} \\
\midrule
count & $n = \sum_{t=1}^T 1$ (number of observed elements) \\
missing & $\mathrm{missing} = \sum_{t=1}^T I(\text{NA}(x_t))$ \\
mean & $\bar{x} = \dfrac{1}{n}\sum_{t=1}^n x_t$ \\
std & $s = \sqrt{\dfrac{1}{n-1}\sum_{t=1}^n (x_t-\bar{x})^2}$ \\
min & $\min_t x_t$ \\
25\% & first quartile $Q_1 = x_{(0.25)}$ (25th percentile) \\
50\%\_median & median $\mathrm{median}=x_{(0.5)}$ (50th percentile) \\
75\% & third quartile $Q_3 = x_{(0.75)}$ (75th percentile) \\
max & $\max_t x_t$ \\
skewness & $\displaystyle \text{skewness} = \frac{\frac{1}{n}\sum_{t=1}^n (x_t-\bar{x})^3}{\left(\frac{1}{n}\sum_{t=1}^n (x_t-\bar{x})^2\right)^{3/2}}$ \\
kurtosis & $\displaystyle \text{kurtosis} = \frac{\frac{1}{n}\sum_{t=1}^n (x_t-\bar{x})^4}{\left(\frac{1}{n}\sum_{t=1}^n (x_t-\bar{x})^2\right)^{2}} - 3$ (excess kurtosis) \\
range & $\mathrm{range} = \max_t x_t - \min_t x_t$ \\
iqr & $\mathrm{IQR} = Q_3 - Q_1$ \\
variance & $\mathrm{Var}(x) = s^2$ \\
acf\_lag\_{k} & $\displaystyle \rho_k = \frac{\sum_{t=k+1}^T (x_t-\bar{x})(x_{t-k}-\bar{x})}{\sum_{t=1}^T (x_t-\bar{x})^2},\quad k\in\{1,3,6,12,24\}$ \\
pacf\_lag\_{k} & Partial autocorrelation at lag $k$, $\phi_{kk}$ from the Yule--Walker or Levinson--Durbin equations (PACF values at $k\in\{1,3,6,12,24\}$). \\
trend\_strength & Fit linear regression $x_t = a + b t + e_t$, then $\text{trend\_strength}=|b|$ (absolute slope) \\
adf\_stat & ADF test regression: $\Delta x_t = \alpha + \beta t + \gamma x_{t-1} + \sum_{i=1}^p \delta_i \Delta x_{t-i} + e_t$. \text{ADF statistic} = $t$-statistic of $\hat{\gamma}$. \\
adf\_pvalue & p-value associated with the ADF statistic (computed via the Dickey--Fuller distribution). \\
num\_peaks & $\sum_{t=2}^{T-1} I(x_{t-1}<x_t \wedge x_t>x_{t+1})$ (local maxima count) \\
num\_troughs & $\sum_{t=2}^{T-1} I(x_{t-1}>x_t \wedge x_t<x_{t+1})$ (local minima count) \\
zero\_ratio & $\dfrac{1}{n}\sum_{t=1}^n I(x_t=0)$ \\
outlier\_ratio & Fraction outside Tukey fences: $\dfrac{1}{n}\sum_{t=1}^n I\big(x_t < Q_1-1.5\cdot\mathrm{IQR} \ \text{or} \ x_t > Q_3+1.5\cdot\mathrm{IQR}\big)$ \\
coef\_of\_variation & $\mathrm{CV} = \dfrac{s}{|\bar{x}|}$ \\
nonlinearity\_proxy & $\mathrm{std}(\Delta x_t) = \sqrt{\mathrm{Var}(x_t-x_{t-1})}$ (std of first differences) \\
trend\_strength\_decomp & From decomposition $x_t = T_t + S_t + R_t$: $\mathrm{trend\_strength\_decomp}=\dfrac{\mathrm{Var}(T_t)}{\mathrm{Var}(x_t)}$ \\
seasonal\_strength\_decomp & $\mathrm{seasonal\_strength\_decomp}=\dfrac{\mathrm{Var}(S_t)}{\mathrm{Var}(x_t)}$ \\
residual\_strength & $\mathrm{residual\_strength}=\dfrac{\mathrm{Var}(R_t)}{\mathrm{Var}(x_t)}$ \\
\bottomrule
\end{tabular}
\end{table}
\paragraph{Model Meta-Knowledge}
A detailed architectural description of the Bi-LSTM is
included, specifying bidirectional processing, stacked layers, dropout usage, loss
function, and optimization objective.
\paragraph{Optimization History}
The best-performing configurations identified by Bayesian
Optimization are provided as historical trials (Table~\ref{tab:bo_trials}). These trials serve as
task-specific meta-knowledge, enabling the LLM to refine strong initial solutions
rather than exploring from scratch.

\begin{table}[h!]
\centering
\small
\caption{Historical Bayesian Optimization trials used as contextual guidance for LLM-based hyperparameter recommendation.}
\label{tab:bo_trials}
\rowcolors{2}{gray!10}{white}
\begin{tabular}{p{1.2cm} p{1cm} p{1.4cm} p{0.8cm} p{1cm} p{1.2cm} p{1.1cm} p{1cm} p{1.4cm}}
\toprule
\textbf{Trial ID} & \textbf{Lag} & \textbf{Hidden} & \textbf{Layers} & \textbf{Dropout} & \textbf{LR} & \textbf{Batch} & \textbf{Epochs} & \textbf{Optimizer} \\
\midrule
trial\_3 & 57 & 56 & 1 & 0.50 & 9.84e-3 & 32 & 12 & Adamax \\
trial\_4 & 46 & 118 & 1 & 0.43 & 3.58e-3 & 64 & 19 & AdamW \\
trial\_1 & 36 & 39 & 1 & 0.27 & 8.44e-4 & 128 & 10 & AdamW \\
\bottomrule
\end{tabular}
\end{table}

\subsubsection{Evaluation Metrics}

We evaluate performance using three complementary metrics designed to assess both solution quality and computational efficiency:

\paragraph{Forecasting Accuracy (RMSE)}
Root Mean Square Error (RMSE) is used to assess predictive accuracy. It is defined as:
\begin{equation}
\text{RMSE} = \sqrt{\frac{1}{N}\sum_{i=1}^{N} (y_i - \hat{y}_i)^2}
\end{equation}
where $y_i$ is the true value at step $i$ and $\hat{y}_i$ is the predicted value, with $N$ being the total number of test samples. RMSE penalizes large errors quadratically and is well suited for continuous forecasting tasks. RMSE serves both as the evaluation metric and as feedback signal for the LLM-guided optimization loop, allowing the framework to iteratively refine hyperparameter configurations based on validation performance.

\paragraph{Training Time}
Training time $T_{\text{train}}(\lambda)$ measures the computational cost of fitting the model given a fixed hyperparameter configuration:
\begin{equation}
T_{\text{train}}(\lambda) = t_{\text{end}} - t_{\text{start}}
\end{equation}
where $t_{\text{start}}$ and $t_{\text{end}}$ are the wall-clock timestamps marking the beginning and end of model training. This metric captures the downstream computational cost induced by different optimization strategies. Hyperparameters such as batch size, number of epochs, and hidden dimensions directly influence training time, making this metric crucial for evaluating practical efficiency.

\paragraph{Optimization Time}
Optimization time measures the total wall-clock time required to complete the hyperparameter optimization process. We compute optimization time separately for each technique:

\textit{Bayesian Optimization Time:} 
\begin{equation}
T_{\text{opt}}^{\text{BO}} = \sum_{i=1}^{N_{\text{BO}}} \left(T_{\text{acq}}^{(i)} + T_{\text{train}}^{(i)} + T_{\text{eval}}^{(i)}\right)
\end{equation}
where $N_{\text{BO}}$ is the number of BO iterations, $T_{\text{acq}}^{(i)}$ is the time to compute the acquisition function and select the next candidate, $T_{\text{train}}^{(i)}$ is the training time for configuration $i$, and $T_{\text{eval}}^{(i)}$ is the validation evaluation time.

\textit{LLM-Guided Optimization Time:}
\begin{equation}
T_{\text{opt}}^{\text{LLM}} = \sum_{t=1}^{M} \left(T_{\text{LLM}}^{(t)} + T_{\text{train}}^{(t)} + T_{\text{eval}}^{(t)}\right)
\end{equation}
where $M$ is the number of LLM-guided refinement iterations, $T_{\text{LLM}}^{(t)}$ is the time for the LLM to generate a hyperparameter recommendation at iteration $t$, $T_{\text{train}}^{(t)}$ is the training time for the recommended configuration, and $T_{\text{eval}}^{(t)}$ is the validation evaluation time.

Computing each technique separately enables direct comparison of computational efficiency between Bayesian Optimization and LLM-guided optimization, facilitating assessment of the trade-off between recommendation quality and computational cost.
\subsection{Results and Discussion}
This section presents a quantitative and qualitative evaluation of the proposed
LLM-AutoOpt framework. Results are organized around the research questions
introduced in Section 4, with emphasis on predictive performance, optimization
efficiency, convergence behavior, and interpretability.

\subsubsection{RQ1 (Performance)}
\begin{figure}[h!]
    \centering
    \includegraphics[width=0.8\textwidth]{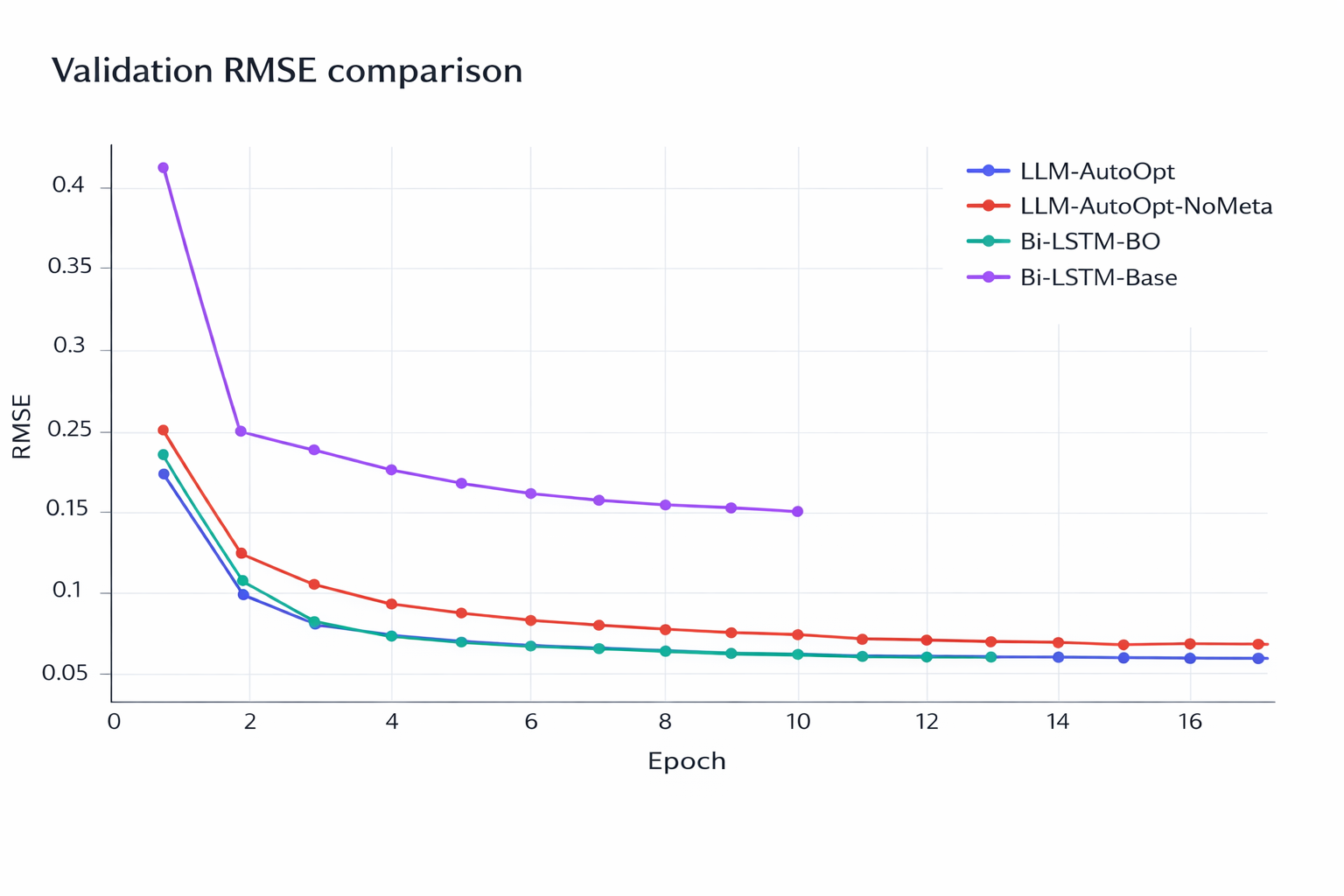}
    \caption{\large Validation RMSE convergence across optimization strategies. LLM-AutoOpt (blue) achieves the lowest final RMSE and fastest convergence compared to BO-only (orange) and LLM without meta-knowledge (green) approaches.}
    \label{fig:rmse_convergence}
\end{figure}

Figure~\ref{fig:rmse_convergence} and Table~\ref{tab:results_performance} compare forecasting accuracy across all optimization strategies using RMSE as the primary metric. 
LLM-AutoOpt consistently achieves the lowest RMSE among all baselines, including Bayesian Optimization (Bi-LSTM-BO), an LLM without meta-knowledge (LLM-AutoOpt-NoMeta), and the manually initialized baseline model (Bi-LSTM-Base). 
LLM-AutoOpt exhibits faster convergence and stabilizes at a significantly lower error level throughout training. 
In contrast, Bi-LSTM-BO converges more slowly and fails to reach the predefined target RMSE, remaining approximately $0.09$ above the threshold.
The LLM-AutoOpt-NoMeta improves over the baseline but demonstrates higher variance and inferior final accuracy,
highlighting the limitations of unguided LLM reasoning.

\begin{table}[h!]
\centering
\caption{Comparison of Mean RMSE, Average Training Time, and Optimization Time for different methods (with standard deviations).}
\label{tab:results_performance}
\rowcolors{2}{gray!10}{white}
\begin{tabular}{p{3.5cm} p{1.8cm} p{1.8cm} p{2.2cm} p{2.2cm}}
\toprule
\textbf{Method} & \textbf{Mean RMSE} & \textbf{RMSE Std} & \textbf{Training Time (s)} & \textbf{Optimization Time (s)} \\
\midrule
Bi-LSTM-Base & 5.42 & 0.31 & 31 & --- \\
Bi-LSTM-BO (Top 3) & 1.19 & 0.03 & 30 & 82 \\
LLM-AutoOpt-NoMeta (Top 3) & 1.90 & 0.05 & 549 & 1065 \\
LLM-AutoOpt (Top 3) & 1.11 & 0.04 & 46 & 177 \\
\bottomrule
\end{tabular}
\end{table}

\begin{figure}[h!]
    \centering
    \includegraphics[width=0.8\textwidth]{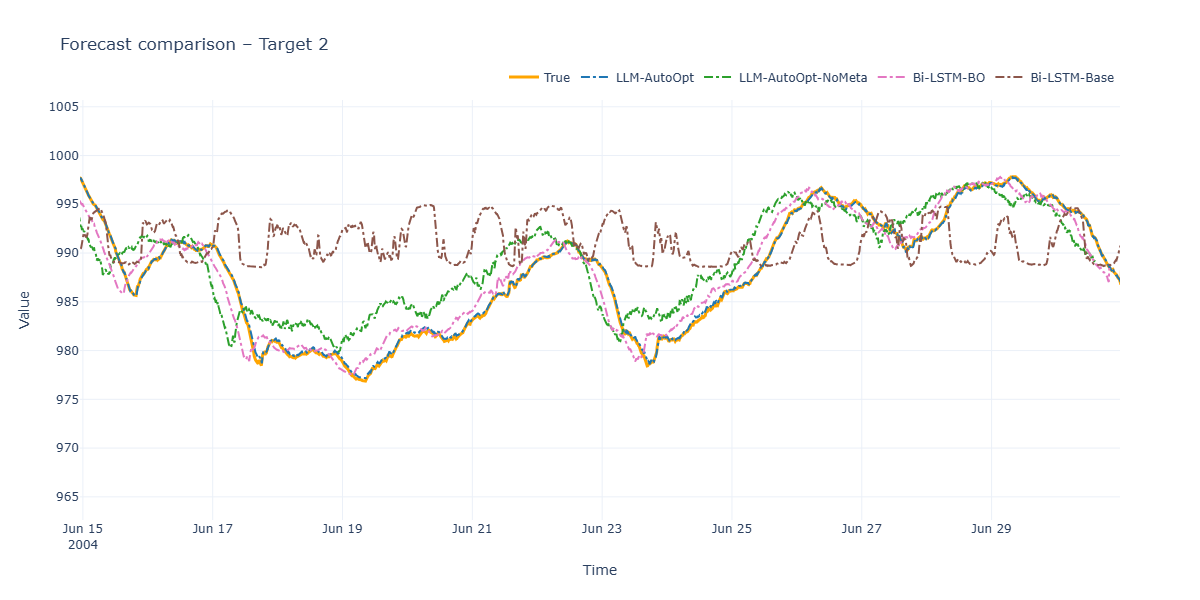}
    \caption{Relative humidity (RH\,\%) forecast comparison across different models}

    \label{plot2:Forecast_results}
\end{figure}

Beyond aggregate error metrics, 
Figure~\ref{plot2:Forecast_results} provides a qualitative comparison of the relative humidity (RH\,\%) forecasts produced by different models. 
The forecasts generated by LLM-AutoOpt more closely track the ground-truth signal, capturing both local fluctuations and longer-term temporal trends.
Competing methods exhibit noticeable smoothing effects or delayed responses to rapid changes,
indicative of suboptimal hyperparameter configurations.

Taken together, these quantitative and qualitative results demonstrate that meta-knowledge–guided LLM 
optimization not only reduces overall forecasting error but also improves the fidelity of 
time-series predictions,particularly in capturing dynamic temporal patterns.

\subsubsection{RQ2: Computational Efficiency}
Table~\ref{tab:results_performance} reports both training time and total optimization time.
Bayesian Optimization remains the most computationally efficient approach, requiring the least optimization time and minimal per-trial training cost.
In contrast, LLM-based methods incur additional overhead due to prompt construction, reasoning, and iterative feedback.
Notably, LLM-AutoOpt is significantly more efficient than the LLM-AutoOpt-NoMeta variant, reducing optimization time by more than $80\%$ while achieving substantially better accuracy. 
This indicates that meta-knowledge not only improves solution quality but also constrains the search space effectively, reducing unnecessary exploration.
While LLM-AutoOpt does not outperform Bi-LSTM-BO in raw computational speed, it offers a different trade-off: improved accuracy and interpretability at the cost of moderate additional overhead.

\subsubsection{RQ3: Optimization Dynamics and Convergence Behavior}

The RMSE trajectories in Figure~\ref{fig:rmse_convergence} reveal distinct optimization dynamics. 
LLM-AutoOpt exhibits fast early-stage improvement followed by stable convergence, 
suggesting that the LLM leverages meta-knowledge to make informed, high-impact adjustments early in the optimization process.
In contrast, the LLM-AutoOpt-NoMeta shows erratic convergence behavior, often relying on brute-force strategies such as increasing the number of epochs to reduce RMSE.
Bi-LSTM-BO follows a smoother but slower trajectory and appears prone to local stagnation near suboptimal regions.
These observations suggest that LLM-guided refinement, when grounded in meta-knowledge, can escape limitations of purely surrogate-based optimization and converge with fewer ineffective trials.
\subsubsection{RQ4: Impact of Meta-Knowledge on Optimization Quality}

Table~\ref{tab:hyperparam_comparison} highlights qualitative differences in the hyperparameter configurations produced by each method. 
LLM-AutoOpt often proposes configurations that resemble strong BO solutions (e.g., similar batch sizes or layer counts) while introducing targeted refinements such as reducing dropout, adjusting lag length, or lowering epoch counts to mitigate overfitting.

\begin{table*}[ht]
\centering
\small
\caption{Hyperparameter configurations and RMSE results across optimization trials with method comparisons.}
\label{tab:hyperparam_comparison}
\rowcolors{2}{gray!10}{white}
\resizebox{\textwidth}{!}{\begin{tabular}{lcccccccccc}
\toprule
\textbf{Trial} & \textbf{Method} & \textbf{Lag} & \textbf{Hidden Size} & \textbf{Layers} &
\textbf{Dropout} & \textbf{LR} & \textbf{Batch} & \textbf{Epochs} & \textbf{Optimizer} & \textbf{RMSE} \\
\midrule

1 & Bi-LSTM-BO& 36 & 39  & 1 & 0.27 & 8.44e-4 & 128 & 10  & AdamW & 1.42 \\
  & LLM-AutoOpt-NoMeta & 24 & 50  & 3 & 0.20 & 1.00e-3 & 64  & 100 & Adam  & 2.10 \\
  & LLM-AutoOpt & 12 & 64  & 2 & 0.15 & 1.00e-3 & 64  & 40  & Adam  & \textbf{1.15} \\
\midrule

2 & Bi-LSTM-BO & 57 & 56  & 1 & 0.50 & 9.84e-3 & 32  & 12  & Adamax  & 1.09 \\
  & LLM-AutoOpt-NoMeta & 24 & 100 & 5 & 0.30 & 5.00e-3 & 128 & 200 & RMSprop & 1.95 \\
  & LLM-AutoOpt & 16 & 48  & 2 & 0.10 & 1.00e-3 & 64 & 30  & Adam    & \textbf{1.18} \\
\midrule

3 & Bi-LSTM-BO & 46 & 118 & 1 & 0.43 & 3.58e-3 & 64  & 19  & AdamW  & 1.09 \\
  & LLM-AutoOpt-NoMeta & 24 & 100 & 5 & 0.30 & 5.00e-3 & 128 & 200 & RMSprop & 1.66 \\
  &LLM-AutoOpt & 12 & 64  & 1 & 0.10 & 1.00e-3 & 64  & 40  & Adam    & \textbf{1.01} \\

\bottomrule
\end{tabular}}
\end{table*}

By contrast, the LLM-AutoOpt-NoMeta frequently selects extreme configurations, particularly large epoch counts, as a generic mechanism for RMSE reduction. This behavior indicates a lack of structural understanding of the task and results in inefficient or unstable optimization.
These findings confirm that historical trials and model-aware context function as actionable meta-knowledge, enabling the LLM to prioritize meaningful adjustments rather than arbitrary exploration.
\subsubsection{RQ5: Interpretability and Reasoning Transparency}
\begin{figure*}[ht]
\centering
\includegraphics[width=0.7\textwidth]{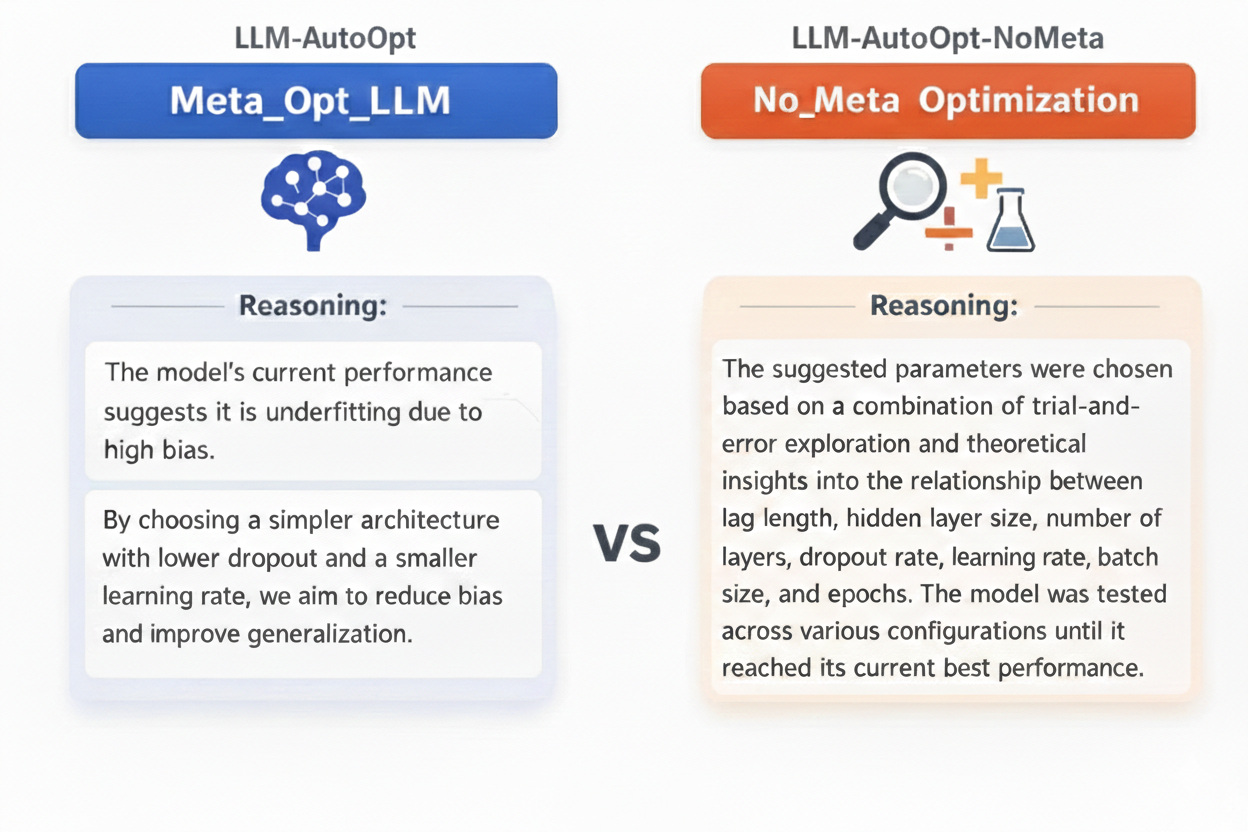}
\caption{Comparison of reasoning outputs for hyperparameter recommendation with and without meta-knowledge injection.}
\label{fig:llm_reasoning_no_meta}
\end{figure*}

Figure~\ref{fig:llm_reasoning_no_meta} contrasts reasoning traces generated with and without meta-knowledge injection.
The  LLM-AutoOpt-NoMeta produces abstract and generic explanations, typically referencing prior RMSE values without explicit justification tied to model architecture or data properties.
In contrast, LLM-AutoOpt provides grounded and task-aware reasoning, explicitly linking hyperparameter changes to architectural considerations (e.g., reducing dropout to alleviate underfitting) 
or temporal characteristics of the data. This transparency enables practitioners to audit, trust, and potentially reuse optimization insights—an advantage absent in black-box methods such as BO.
Overall, meta-informed reasoning improves not only optimization outcomes but also the interpretability and explanatory power of the optimization process.

Overall, incorporating meta-knowledge into LLM-AutoOpt leads to (i) improved forecasting accuracy and faster convergence compared with manual, Bayesian, and meta-agnostic LLM optimization methods, 
(ii) enhanced interpretability through structured priors and transparent reasoning that connect dataset characteristics to hyperparameter decisions, 
and (iii) a trade-off between performance and computational cost, with effectiveness depending on prompt design and requiring broader evaluation to confirm scalability and generalization.

\section{Conclusion}\label{sec5}
This work presented LLM-AutoOpt, a meta-knowledge–guided hyperparameter optimization framework that leverages large language model reasoning for multivariate time-series forecasting. By integrating dataset meta-features, model descriptions, and historical optimization results into a structured prompting strategy, the framework enables more accurate, stable, and interpretable hyperparameter tuning.
Empirical results show that LLM-AutoOpt consistently outperforms traditional Bayesian Optimization and LLM-based baselines without meta-knowledge in terms of predictive accuracy, while providing transparent and data-aware optimization reasoning. Although the approach does not improve computational efficiency, it offers a principled and explainable alternative for high-stakes forecasting tasks.

Future work will explore cross-dataset generalization, automated meta-feature selection, and extensions to additional forecasting domains and deep sequence models, further advancing the role of meta-knowledge–augmented LLMs in hyperparameter optimization.

\bibliography{sn-bibliography}

@online{ibm_automl_2025,
  author  = {IBM},
  title   = {What is AutoML?},
  year    = {2025},
  url     = {https://www.ibm.com/think/topics/automl},
  urldate = {2025-12-03}
}

@article{bib1,
  author  = {Franceschi, Luca and Donini, Michele and Perrone, Valerio and
             Klein, Aaron and Archambeau, Cédric and Seeger, Matthias and
             Pontil, Massimiliano and Frasconi, Paolo},
  title   = {Hyperparameter Optimization in Machine Learning},
  journal = {arXiv preprint arXiv:2410.22854},
  year    = {2024}
}

@article{bib2,
  author  = {Andonie, Răzvan},
  title   = {Hyperparameter Optimization in Learning Systems},
  journal = {Journal of Membrane Computing},
  volume  = {1},
  number  = {4},
  pages   = {279--291},
  year    = {2019}
}

@article{bib3,
  author  = {Bischl, Bernd and Binder, Martin and Lang, Michel and Pielok, Tobias and
             Richter, Jakob and Coors, Stefan and Thomas, Janek and Ullmann, Tim and
             Becker, Marc and Boulesteix, Anne-Laure and others},
  title   = {Hyperparameter Optimization: Foundations, Algorithms, Best Practices and Open Challenges},
  journal = {arXiv preprint arXiv:2107.05847},
  year    = {2021}
}

@article{bib4,
  author  = {Rivolli, Adriano and Garcia, Luís P. F. and Soares, Carlos and others},
  title   = {Meta-features for Meta-learning},
  journal = {Knowledge-Based Systems},
  volume  = {240},
  pages   = {108101},
  year    = {2022}
}

@article{bib5,
  author  = {Rivolli, Adriano and Garcia, Luis P. F. and Soares, Carlos and others},
  title   = {Meta-features for Meta-learning},
  journal = {Knowledge-Based Systems},
  volume  = {240},
  pages   = {108101},
  year    = {2022}
}

@article{bib6,
  author  = {Rivolli, Adriano and Garcia, Luis P. F. and Soares, Carlos and others},
  title   = {Meta-features for Meta-learning},
  journal = {Knowledge-Based Systems},
  volume  = {240},
  pages   = {108101},
  year    = {2022}
}

@article{bib7,
  author  = {Vanschoren, Joaquin},
  title   = {Meta-learning: A Survey},
  journal = {arXiv preprint arXiv:1810.03548},
  year    = {2018}
}

@inproceedings{bib8,
  author    = {Wistuba, Martin and Schilling, Nico and Schmidt-Thieme, Lars},
  title     = {Sequential Model-Free Hyperparameter Tuning},
  booktitle = {Proceedings of the IEEE International Conference on Data Mining (ICDM)},
  pages     = {1033--1038},
  year      = {2015},
  doi       = {10.1109/ICDM.2015.20}
}

@inproceedings{bib9,
  author    = {Feurer, Matthias and Springenberg, Jost Tobias and Hutter, Frank},
  title     = {Initializing Bayesian Hyperparameter Optimization via Meta-learning},
  booktitle = {Proceedings of the AAAI Conference on Artificial Intelligence (AAAI)},
  pages     = {1128--1135},
  year      = {2015}
}

@inproceedings{bib10,
  author    = {Lindauer, Marius and Hutter, Frank},
  title     = {Warmstarting of Model-Based Algorithm Configuration},
  booktitle = {Proceedings of the AAAI Conference on Artificial Intelligence (AAAI-18)},
  pages     = {1355--1362},
  year      = {2018}
}

@article{bib11,
  author  = {Feurer, Matthias and Eggensperger, Katharina and Falkner, Stefan and
             Lindauer, Marius and Hutter, Frank},
  title   = {Auto-sklearn 2.0: Hands-free AutoML via Meta-learning},
  journal = {Journal of Machine Learning Research},
  volume  = {23},
  number  = {261},
  pages   = {1--61},
  year    = {2022}
}

@article{bib12,
  author  = {Zheng, Mingkai and Su, Xiu and You, Shan and Wang, Fei and
             Qian, Chen and Xu, Chang and Albanie, Samuel},
  title   = {Can {GPT}-4 Perform Neural Architecture Search?},
  journal = {arXiv preprint arXiv:2304.10970},
  year    = {2023}
}

@article{bib13,
  author  = {Yang, Chengrun and Wang, Xuezhi and Lu, Yifeng and Liu, Hanxiao and
             Le, Quoc V. and Zhou, Denny and Chen, Xinyun},
  title   = {Large Language Models as Optimizers},
  journal = {arXiv preprint arXiv:2309.03409},
  year    = {2023}
}

@inproceedings{bib14,
  author    = {Brown, Tom and Mann, Benjamin and Ryder, Nick and Subbiah, Melanie and
               Kaplan, Jared D. and Dhariwal, Prafulla and Neelakantan, Arvind and
               Shyam, Pranav and Sastry, Girish and Askell, Amanda and others},
  title     = {Language Models are Few-Shot Learners},
  booktitle = {Advances in Neural Information Processing Systems (NeurIPS)},
  volume    = {33},
  pages     = {1877--1901},
  year      = {2020}
}

@article{bib15,
  author  = {Zhao, Wayne Xin and Zhou, Kun and Li, Junyi and Tang, Tianyi and
             Wang, Xiaolei and Hou, Yupeng and Min, Yingqian and Zhang, Beichen and
             Zhang, Junjie and Dong, Zican and others},
  title   = {A Survey of Large Language Models},
  journal = {arXiv preprint arXiv:2303.18223},
  year    = {2023}
}

@article{bib16,
  author  = {Liu, Tennison and Astorga, Nicol{\'a}s and Seedat, Nabeel and others},
  title   = {Large Language Models to Enhance Bayesian Optimization},
  journal = {arXiv preprint arXiv:2402.03921},
  year    = {2024}
}

@article{bib17,
  author  = {Hili, Youssef Attia El and others},
  title   = {LLMs as In-Context Meta-Learners for Model and Hyperparameter Selection},
  journal = {arXiv preprint arXiv:2510.26510},
  year    = {2025}
}

@article{bib18,
  author  = {Bal-Ghaoui, Mohamed and Tiouti, Mohammed},
  title   = {MetaLLMix: An XAI Aided LLM-Meta-learning Based Approach for Hyper-parameters Optimization},
  journal = {arXiv preprint arXiv:2509.09387},
  year    = {2025}
}

@misc{bib19,
  author = {Dimgba, Martha Otisi and Andreas, R.},
  title  = {Optimal Hyperparameter Search Strategies: Benchmarking Grid Search, Random Search, and Genetic Algorithms across Regression, Classification, and Clustering Tasks},
  year   = {2024}
}

@article{bib20,
  author  = {Bergstra, James and Bengio, Yoshua},
  title   = {Random Search for Hyper-Parameter Optimization},
  journal = {Journal of Machine Learning Research},
  volume  = {13},
  number  = {1},
  pages   = {281--305},
  year    = {2012}
}

@misc{bib21,
  author       = {{Wikipedia contributors}},
  title        = {Hyperparameter Optimization --- {Wikipedia}{,} The Free Encyclopedia},
  year         = {2025},
  howpublished = {\url{https://en.wikipedia.org/wiki/Hyperparameter_optimization}},
  note         = {Accessed: 12-Feb-2026}
}

@article{bib22,
  author  = {Vettoruzzo, Anna and others},
  title   = {Advances and Challenges in Meta-Learning: A Technical Review},
  journal = {IEEE Transactions on Pattern Analysis and Machine Intelligence},
  volume  = {46},
  number  = {7},
  pages   = {4763--4779},
  year    = {2024}
}

@article{bib23,
  author  = {Wang, H. and Sun, Z. and Du, Y. and Zhang, L. and He, T. and Ong, Y. S.},
  title   = {Uncertain Multi-Objective Recommendation via Orthogonal Meta Learning Enhanced Bayesian Optimization},
  journal = {arXiv preprint arXiv:2502.13180},
  year    = {2025}
}

@article{bib24,
  author  = {Tunio, Muhammad Hanif and Li, Jian Ping and Zeng, Xiaoyang and Akhtar, Faijan and
             Shah, Syed Attique and Ahmed, Awais and Yang, Yu and Heyat, Md Belal Bin},
  title   = {Meta-knowledge Guided Bayesian Optimization Framework for Robust Crop Yield Estimation},
  journal = {Journal of King Saud University - Computer and Information Sciences},
  volume  = {36},
  number  = {1},
  pages   = {101895},
  year    = {2024}
}

@article{bib25,
  author  = {Iqbal, Muhammad Ali and Gil, Joon-Min and Kim, Soo Kyun},
  title   = {Attention-driven Hybrid Ensemble Approach with Bayesian Optimization for Accurate Energy Forecasting in Jeju Island’s Renewable Energy System},
  journal = {IEEE Access},
  year    = {2025}
}

@article{bib26,
  author  = {Balef, Amir Rezaei and Vernade, Claire and Eggensperger, Katharina},
  title   = {Put CASH on Bandits: A Max K-Armed Problem for Automated Machine Learning},
  journal = {arXiv preprint arXiv:2505.05226},
  year    = {2025}
}

@misc{bib27,
  author  = {Julian Optimization In V. and Camacho, D. and Yin, H. and Alberola, J. M. and
             Nogueira, V. B. and Novais, P. and Tallón-Ballesteros, A.},
  title   = {Model-Based Meta-reinforcement Learning for Hyperparameter Optimization},
  year    = {2025}
}

@article{bib28,
  author  = {Hog, Johannes and Rajan, Raghu and Biedenkapp, André and Awad, Noor and
             Hutter, Frank and Nguyen, Vu},
  title   = {Meta-learning Population-based Methods for Reinforcement Learning},
  journal = {Transactions on Machine Learning Research},
  year    = {2025}
}

@article{bib29,
  author  = {Liu, Siyi and Gao, Chen and Li, Yong and others},
  title   = {Large Language Model Agent for Hyper-parameter Optimization},
  journal = {arXiv preprint arXiv:2402.01881},
  year    = {2024}
}

@article{bib30,
  author  = {Zhang, Michael R. and Desai, Nishkrit and Bae, Juhan and Lorraine, Jonathan and Ba, Jimmy},
  title   = {Using Large Language Models for Hyperparameter Optimization},
  journal = {arXiv preprint arXiv:2312.04528},
  year    = {2023}
}

@inproceedings{bib32,
  author    = {Kalele, Nupur and Kalyani, Vijay Kumar},
  title     = {Grid Search and LLM Assisted Hybrid Approach for Hyperparameter Optimization},
  booktitle = {2025 IEEE International Conference on Interdisciplinary Approaches in Technology and Management for Social Innovation (IATMSI)},
  volume    = {3},
  publisher = {IEEE},
  year      = {2025}
}

@misc{bib33,
  author  = {Kroeger, Nicholas and Ley, Dan and Krishna, Satyapriya and Agarwal, Chirag and Lakkaraju, Himabindu},
  title   = {In-Context Explainers: Harnessing LLMs for Explaining Black Box Models},
  year    = {2024}
}

@misc{bib34,
  author       = {Mnassrib, M.},
  title        = {Jena Climate Dataset},
  year         = {2025},
  howpublished = {\url{https://www.kaggle.com/datasets/mnassrib/jena-weather-dataset}},
  note         = {Accessed: November 26, 2025}
}

@article{bib35,
  author  = {Ihianle, Isibor Kennedy and Nwajana, Augustine O. and Ebenuwa, Solomon Henry and Otuka, Richard I. and Owa, Kayode and Orisatoki, Mobolaji O.},
  title   = {A Deep Learning Approach for Human Activities Recognition from Multimodal Sensing Devices},
  journal = {IEEE Access},
  volume  = {8},
  pages   = {179028--179038},
  year    = {2020}
}

\end{document}